# PSO based Neural Networks vs. Traditional Statistical Models for Seasonal Time Series Forecasting

Ratnadip Adhikari
Computer and Systems Sciences
Jawaharlal Nehru University
New Delhi, India
Email: adhikari.ratan@gmail.com

R. K. Agrawal
Computer and Systems Sciences
Jawaharlal Nehru University
New Delhi, India
Email: rkajnu@gmail.com

Laxmi Kant
Department of Mathematics
IIT Roorkee
Roorkee, India
Email: laxmikant75@gmail.com

*Abstract*—Seasonality is a distinctive characteristic which is often observed in many practical time series. Artificial Neural Networks (ANNs) are a class of promising models for efficiently recognizing and forecasting seasonal patterns. In this paper, the Particle Swarm Optimization (PSO) approach is used to enhance the forecasting strengths of feedforward ANN (FANN) as well as Elman ANN (EANN) models for seasonal data. Three widely popular versions of the basic PSO algorithm, viz. Trelea-I, Trelea-II and Clerc-Type1 are considered here. The empirical analysis is conducted on three real-world seasonal time series. Results clearly show that each version of the PSO algorithm achieves notably better forecasting accuracies than the standard Backpropagation (BP) training method for both FANN and EANN models. The neural network forecasting results are also compared with those from the three traditional statistical models, viz. Seasonal Autoregressive Integrated Moving Average (SARIMA), Holt-Winters (HW) and Support Vector Machine (SVM). The comparison demonstrates that both PSO and BP based neural networks outperform SARIMA, HW and SVM models for all three time series datasets. The forecasting performances of ANNs are further improved through combining the outputs from the three PSO based models.

## I. INTRODUCTION

Trends and seasonal effects are frequently observed in many time series, especially those pertaining to economics, business, finance and natural phenomena. Seasonality is a kind of regular repetitive fluctuation which occurs within a year, often in a quarterly or monthly basis. The seasonal patterns introduce additional intricacies to a time series, thereby making the task of apposite modeling and forecasting reasonably difficult [1]. To date, there are a few recognized approaches to systematically analyze and forecast seasonal data. Some popular old methods include the moving average [2], exponential smoothing [2], Holt-Winters (HW) [3], etc. Over the past four decades, the Seasonal Autoregressive Integrated Moving Average (SARIMA) model [4] occupies the leading position among all statistical methods of forecasting seasonal data. More recently, Support Vector Machine (SVM) [3, 5] has also found notable applications in this area. It is imperative to know that most of these traditional methods were developed decades ago and they adopt a rather fixed model for a seasonal time series [3]. This restricts their flexibility through making them inadequate to cope with the frequent dynamic changes in seasonal patterns. Furthermore, the traditional models solely require the pre-removal of the seasonal effect from a time series through techniques, such as *seasonal differencing* and *deseasonalization* before making future forecasts. Such techniques often distort many fundamental properties of the time series and so they have been criticized by numerous well respected researchers [1, 6, 7].

It should not be surprising that Artificial Neural Networks (ANNs) have gained massive popularity in modeling seasonal data. ANNs have already become a research hotspot in time series analysis and forecasting due to their several amazing characteristics, a comprehensive study of which can be found in the work of Zhang *et al.* [8]. A number of prominent researchers, such as Alon *et al.* [9], Tseng *et al.* [10], Kihoro and Otieno [11] observed that ANNs have excellent ability in properly recognizing and forecasting the seasonal fluctuations. However, their findings were also disagreed by an equally large number of researchers. For example, according to Faraway and Chatfield [12], Hill *et al.* [13], Zhang and Qi [14], ANNs are not appropriate for seasonal time series forecasting. Despite these earlier contradictory conclusions, many recent studies strongly support that ANNs are superior to the traditional statistical models in forecasting seasonal data [3, 15, 16].

Over the past few years, evolutionary computing techniques have attracted considerable attention of artificial intelligence researchers. One of such techniques is the Particle swarm Optimization (PSO), developed by Kennedy and Eberhart [17]. PSO provides an intelligent mathematical formulation of the working mechanism of birds flock in order to iteratively optimize nonlinear multidimensional functions. It has high search power in the state space, fast convergence rate and ability to provide global optimal solution [18, 19]. Due to these salient properties, PSO has been effectively used by some researchers as an alternative to the Backpropagation (BP) algorithm for training ANN models [19, 20]. PSO overcomes many inherent drawbacks of the BP algorithm which include large computational time, slow convergence rate, getting stuck at local minima, etc. [19]. However, PSO based ANN models have not yet found reasonable number of applications in forecasting time series with trends or seasonal patterns.

In this paper, two common neural network structures, viz. feedforward ANN (FANN) and recurrent ANN of Elman Type

(EANN) are trained through three widely popular variants of the PSO algorithm in order to forecast seasonal time series data. These three variants of PSO are PSO Trelea-I [20, 21], PSO Trelea-II [20, 21] and PSO Clerc-Type1 [22]. Empirical analysis is conducted on three real-world seasonal time series. The forecasting accuracies of the PSO based ANNs are compared with the corresponding BP based ANNs as well as the three traditional statistical methods, viz. SARIMA, HW and SVM for all three seasonal time series.

The rest of the paper is organized as follows. Section II and Section III respectively describes the three traditional statistical methods and ANNs for seasonal time series forecasting. Section IV describes the PSO algorithm with its three variants. The empirical findings are reported in Section V and finally the paper is concluded in Section VI.

## II. TRADITIONAL SEASONALITY FORECASTING METHODS

The identification of the nature and period of seasonal variation is crucial for appropriate modeling and forecasting. In many cases, the *correlogram* of the time series provides a fairly good idea about its seasonal structure [3, 4]. A correlogram of the time series $\{y_1, y_2, \ldots, y_N\}$ with mean $\bar{y}$ is a plot of the sample autocorrelation coefficients $r_k$ against the successive time lags $k$, where:

$$r_k = \frac{\sum_{t=1}^{N-k}(y_t - \bar{y})(y_{t+k} - \bar{y})}{\sum_{t=1}^{N}(y_t - \bar{y})^2} \quad (1)$$

$$\forall k = 0, 1, \ldots, (N-1).$$

The correlogram of a seasonal time series shows the same kind of oscillations at the seasonal lags and so can be used to manually detect the seasonal patterns. Other more robust automatic seasonality detection tools are also used. Recently, Yan [23] employed a simple rule of thumb for identifying seasonality. According to him, a time series of length $N$ has a seasonality of period $s$ only if

$$\left. \begin{array}{l} r_s > \frac{2}{\sqrt{N}} \; (\text{for } N \leq 60) \\ r_s \text{ and } r_{2s} > \frac{2}{\sqrt{N}} \; (\text{for } N > 60) \end{array} \right\} \quad (2)$$

Through this criterion, Yan [23] could correctly identify 45 out of all 54 seasonal series and 53 out of all 57 nonseasonal series of the NN3 competition data. Next, we are going to discuss about the three well recognized traditional statistical models for seasonal time series forecasting.

### A. Box-Jenkins Model

The most well-known statistical technique for seasonal time series forecasting is the Seasonal autoregressive Integrated Moving average (SARIMA) model which is also commonly known as the Box-Jenkins model [4, 10, 16]. Mathematically, a SARIMA$(p,d,q) \times (P,D,Q)^s$ model is given by:

$$\phi_p(B)\Phi_P(B^s)W_t = \theta_q(B)\Theta_Q(B^s)Z_t \quad (3)$$

where, $s$ is the period of seasonality; $B$ is the backshift operator, defined as $By_t = y_{t-1}$; $\phi_p, \Phi_P, \theta_q, \Theta_Q$ are the lagged polynomials in $B$ of orders $p$, $P$, $q$ and $Q$ respectively; $Z_t$ is a series of random errors and $W_t$ is the stationary, nonseasonal series which is obtained after the ordinary and seasonal differencing processes, i.e.

$$W_t = (1-B)^d (1-B^s)^D y_t \quad (4)$$

The SARIMA model transforms a seasonal time series to a stationary nonseasonal one through applying ordinary and seasonal differencing sequences to the series [4, 16]. The most appropriate SARIMA model for a particular forecasting problem is usually determined through the famous Box-Jenkins three step iterative model building methodology [4].

### B. Holt-Winters (HW) Model

The Holt-Winters (HW) model belongs to the family of exponential smoothing techniques. The multiplicative HW model iteratively updates the *local mean level* ($L_t$), *trend* ($T_t$), and *seasonal index* ($I_t$) of a time series as follows [3, 24]:

$$\left. \begin{array}{l} L_t = \alpha(y_t / I_{t-s}) + (1-\alpha)(L_{t-1} + I_{t-1}) \\ T_t = \gamma(L_t - L_{t-1}) + (1-\gamma)T_{t-1} \\ I_t = \delta(y_t / L_t) + (1-\delta)I_{t-s} \end{array} \right\} \quad (5)$$

$$\forall t = 0, 1, 2, \ldots, N$$

where, $\alpha$, $\gamma$, and $\delta$ are the smoothing parameters and $s$ is the seasonal period. Using the multiplicative HW model, the $k$-step ahead forecast of $y(t)$ is given by:

$$\hat{y}(t) = (L_t + kT_t)I_{t-s+k} \quad (6)$$

$$\forall k = 1, 2, \ldots, s.$$

There are also analogous formulae for the additive case. Chatfield and Yar [24] presented a detailed study about the fitting of HW model and in this paper we precisely follow their guidelines and recommendations.

### C. Support Vector Machine (SVM) Model

During the past few years, SVMs [25] have found notable applications in the domain of time series modeling and forecasting. These are based on the Structural Risk Minimization (SRM) principle with the objective to find a linear decision rule with good generalization ability. SVM maps the input space into a higher dimensional feature space by using kernel functions [3, 5, 25].

A time series forecasting problem with $N$ training pairs $\{\mathbf{x}_i, y_i\}_{i=1}^{N}$, $\mathbf{x}_i \in R^n, y_i \in R$ falls in the category of Support Vector Regression (SVR) whose aim is to find a maximum margin hyperplane in order to classify real-valued outputs. Using Vapnik's $\varepsilon$-insensitive loss function [3, 25], the SVR is converted to a Quadratic Programming Problem (QPP) in order to minimize the empirical risk:

$$J(\mathbf{w}, \xi_i, \xi_i^*) = \frac{1}{2}\|\mathbf{w}\|^2 + C\sum_{i=1}^{N}(\xi_i + \xi_i^*) \quad (7)$$

where, $C$ is the positive regularization constant which acts as a penalty to misfit, $\xi_i, \xi_i^*$ are the slack variables and **w** is the weight vector. A solution of the QPP yields the optimal decision hyperplane as follows:

$$y(\mathbf{x}) = \sum_{i=1}^{N_s} (\alpha_i - \alpha_i^*) K(\mathbf{x}, \mathbf{x}_i) + b_{\text{opt}} \qquad (8)$$

where, $N_s$ is the number of support vectors, $\alpha_i, \alpha_i^*$ are the Lagrange multipliers, $b_{\text{opt}}$ is the optimal bias and $K(\mathbf{x}, \mathbf{x}_i)$ is the kernel function. In this paper, we use the Radial Basis function (RBF), defined as $K(\mathbf{x}, \mathbf{y}) = \exp(-\|\mathbf{x}-\mathbf{y}\|^2 / 2\sigma^2)$ as the SVM kernel. The optimal SVM parameters, i.e. $C$ and $\sigma$ are selected through the usual grid search technique [3, 5].

## III. ANN APPROACH FOR TIME SERIES FORECASTING

ANNs are the most widely used computational intelligence models for time series analysis and forecasting. They differ from the traditional statistical forecasting methods due to their data-driven and self-adaptive nature [2, 3, 8]. ANNs can be truly referred as model free structures because they do not need any prior knowledge about the intrinsic data generating process. The appropriate network structure is determined solely on the basis of available input and target patterns. ANNs are also favored due to their distinctive ability of nonlinear modeling with remarkable accuracies.

Different types of neural network structures have been proposed in literature and among them the Feedforward ANN (FANN) is most popular in forecasting applications [2, 8]. Recurrent neural networks have also been recently used in time series forecasting problems, although to limited extents [2, 28]. In this paper, we consider the recurrent ANN model of Elman type (EANN) to forecast seasonal data.

### A. FANN model

A typical FANN consists of many processing units or nodes which are distributed in multiple layers, viz. an *input layer*, one or more *hidden layers* and an *output layer*. The nodes in each layer are connected to those in the immediate next layer through acyclic feedforward connections. A single layer of hidden units are enough to provide the desired accuracy in most of the forecasting situations [8].

In a fully connected FANN model with $p$ input, $h$ hidden and a single output node, the relationship between the inputs $y_{t-i}$ ($i$=1, 2, …, $p$) and the output $y_t$ is given by:

$$y_t = G\left( \alpha_0 + \sum_{j=1}^{h} \alpha_j F\left( \beta_{0j} + \sum_{i=1}^{p} \beta_{ij} y_{t-i} \right) \right) \qquad (9)$$

where, $\alpha_j$, $\beta_{ij}$ ($i=1,2,\ldots,p$; $j=1,2,\ldots,h$) are the connection weights, $\alpha_0$, $\beta_{0j}$ are the bias terms and $F$, $G$ are the network activation functions. The common choice for the activation functions are the *logistic* function for the input layer and the *identity* function for the output layer [2, 8]. In order to ensure a nonlinear input-output mapping, the parameter $h$ is taken to be non-zero. The model, presented in Eq. 9 is commonly referred as a $p \times h \times 1$ FANN model, the architecture of which is diagrammatically depicted in Fig. 1.

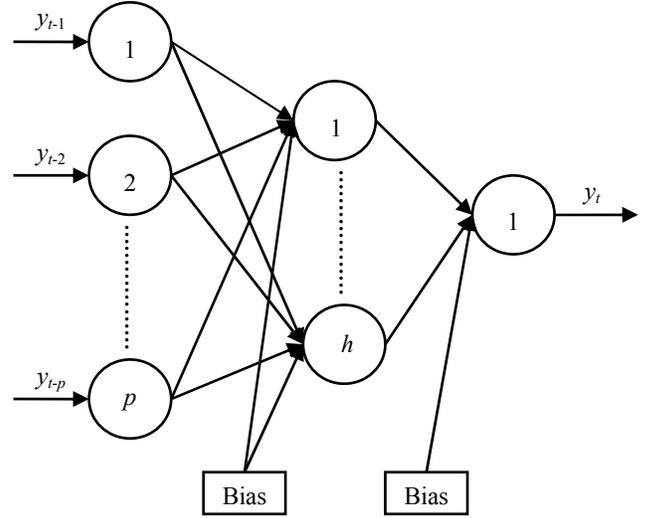

Fig. 1. A $p \times h \times 1$ FANN model with the slant arrows as the network weights

The associated network weights and biases in an ANN model are optimized through the *training* process. The ANN training is an unconstrained nonlinear minimization problem which iteratively updates the network parameters with the goal of minimizing the overall Sum Squared Error (SSE) between the desired and actual output values. The best known training algorithm is the standard backpropagation (BP). It modifies the weights and biases towards the fastest decrease of the error function, i.e. towards the negative of the gradient [8]. In spite of its immense popularity, the standard BP algorithm has some crucial shortcomings which include fairly large number of computations, slow rate of convergence, complex pattern of the error surface, getting trapped into local minima [8, 19, 20]. Although a lot of modifications have been developed in literature, none of them could overcome all the drawbacks of the BP algorithm; for example, none can currently guarantee the global optimal solution [8]. These issues led to the use of evolutionary computation methods for ANN training.

### B. EANN model

Together with the most common feedforward neural networks, other types of network architectures have also been investigated for time series modeling and forecasting. A prominent among them is the Elman ANN (EANN) model. An EANN has a recurrent neural network structure, consisting of a new *context layer* and *feedback* connections [26]. At each step, the outputs from the hidden layer are again fed back to the context layer in order to make the network able to perform dynamical time-varying mappings of the associated nodes. There seems to be no recognized study which firmly establishes the superiority of either FANN or EANN for time series forecasting. However, EANNs are often more robust in adequately modeling the temporal relationships among the input data [2, 26].

### C. Seasonal Time Series Forecasting through ANNs

The traditional statistical methods for seasonal time series forecasting suffer from various drawbacks. These include the assumption of linearity, fixed model form, removal of seasonal effect through deseaonalization or seasonal differencing, etc.

ANNs can overcome many of these limitations and so has found numerous effective applications in modeling and forecasting of seasonal time series. Faraway and Chatfield [12] studied several ANN structures for appropriately modeling the well-known monthly airline data. Alon *et al.* [9] compared the performances of ANN with four traditional methods for predicting US aggregate retail sales which have strong trends and seasonal fluctuations. Zhang and Qi [14] comprehensively investigated the effects of various data preprocessing techniques on neural network forecasting of seasonal time series. More recently, Zhang and Kline [1] studied the forecasting capabilities of 48 different neural network models for a large set of 756 quarterly time series, collected from the M3 competition. Combinations of ANN and other methods were also examined for modeling seasonal data. For example, Tseng *et al.* [10] proposed the SARIMABP model which combines SARIMA and BP-ANN models. Their findings show that the SARIMABP model outperformed the SARIMA, the BP-ANN with deseasonalized data and the BP-ANN with differenced data [10]. Despite the great potential of ANNs in forecasting seasonal data, the earlier studies yielded mixed results. While some find that ANNs can directly model seasonal effects without removing them, others have the just opposite view [9, 10, 12–14].

Very recently, Hamzaçebi [16] has proposed the novel Seasonal ANN (SANN) model in which the number of both input and output nodes is taken to be equal to the seasonal period of the time series, as shown in Fig. 2. From the empirical studies, Hamzaçebi [16] has found that the SANN model can successfully recognize and forecast seasonal effects without removing them from the associated data. It is also quite simple to design and implement because the network structure is already specified and only the number of appropriate hidden nodes is to be determined. In this paper, we use the SANN structure for feedforward as well as Elman networks and respectively refer them as the Seasonal FANN (SFANN) and Seasonal EANN (SEANN) models.

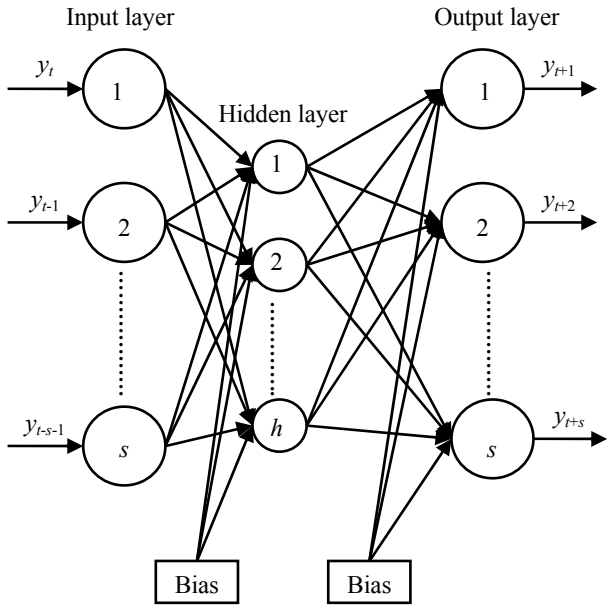

Fig. 2. The structure of a typical $s \times h \times s$ SANN model

*D. Designing of the Appropriate ANN Model*

The successful forecasting through an ANN largely depends on the appropriate model designing which is however not a trivial task [8, 11, 12]. As mentioned earlier, the prime benefit of using SANN is that with this model, the architecture selection actually boils down to the selection of the optimal number of hidden nodes only. In this paper, the number of hidden nodes is selected through the widely popular *Bayesian Information Criterion (BIC)* [11, 12]. For the $s \times h \times s$ SANN model, the BIC is mathematically given by:

$$\text{BIC} = N_{s,h} + N_{s,h} \ln(n) + n \ln\left(\frac{S(\mathbf{W})}{n}\right) \quad (10)$$

where, $N_{s,h} = s + h(2s+1)$ is the number of total network parameters, $n = N - s$ is the number of effective observations, $N$ being the size of the training set, $\mathbf{W}$ is the space of all connection weights and biases and $S(\mathbf{W})$ is the network misfit function which is commonly taken as the SSE. The BIC effectively controls the network size by penalizing for each increase in the number of network parameters [12]. Out of several feedforward SANN structures, the one which minimizes the BIC is chosen to be the optimal one. It should however be noted that the use of BIC is popular in feedforward neural networks only and there seems to be no similar criterion for EANN models. But, it is well-known that EANNs require more hidden nodes than their feedforward counterpart [2]. We use the number of hidden nodes as $2s$ for each EANN model, $s$ being the seasonal period.

The choice of the appropriate training algorithm is another crucial point in ANN model designing. In this paper, we use the *Levenberg-Marquardt (LM)* [27] and *traingdx* [28] as the BP training algorithms for feedforward and recurrent ANN models respectively.

IV. PSO ALGORITHM FOR TRAINING NEURAL NETWORKS

The celebrated PSO algorithm is based on the evolutionary computation meta-heuristic [17, 20]. In this paper, we use PSO for training feedforward as well as recurrent SANN models. Let us consider the $s \times h \times s$ SANN structure, $s$ and $h$ respectively being the period of seasonality and the number of hidden nodes. The PSO algorithm starts with a population (also called *swarm*), consisting of some predefined number of particles which are initialized with random *positions* and *velocities*. If there are $N_{\text{swarm}}$ swarm particles, then each one of them is of dimension $D = s + h(2s+1)$ which is equal to the number of total network parameters. The position of each particle is assigned through evaluating a *fitness function* for it. The particles are moved through the search space on the basis of two best positions, viz. *personal* and *global* best. The personal best position of a particle is the best fitness achieved by it so far, whereas the global best position is the best fitness achieved so far across the whole swarm. Then, the velocity and position of the $i^{\text{th}}$ particle at the $d^{\text{th}}$ dimension are updated as:

$$v_{id}(t+1) = av_{id}(t) + b_1 r_1 \left(p_{id} - x_{id}(t)\right) + b_2 r_2 \left(p_{gd} - x_{id}(t)\right) \quad (11)$$

$$x_{id}(t+1) = x_{id}(t) + v_{id}(t+1) \quad (12)$$

where, $x_{id}$ and $v_{id}$ are respectively the position and velocity of the $i^{th}$ particle at the $d^{th}$ dimension; $p_{id}$ and $p_{gd}$ are respectively the personal and global best positions, achieved so far at the $d^{th}$ dimension; $b_1$ and $b_2$ are the acceleration coefficients; $r_1$ and $r_2$ are two uniform random variables in the [0, 1] interval and $a$ is the inertia weight. The updating process, presented through Eq. 11 and Eq. 12 continues until some predefined stopping criterion, e.g. the maximum number of iterations or the maximum increase in the validation error is attained [19, 20].

In order to improve performances of the PSO algorithm in practical problems, several variants of it have been developed in literature. Among them, the versions of Trelea [21] and Clerc [22] are used in this paper.

*A. PSO Trelea-I and PSO Trelea-II*

Trelea [21] proposed an improved deterministic version of the basic PSO algorithm. The associated formulae for velocity and position updating are given by:

$$v_{id}(t+1) = av_{id}(t) + b(p_d - x_{id}(t)) \quad (13)$$

$$x_{id}(t+1) = x_{id}(t) + v_{id}(t+1) \quad (14)$$

where,

$$r_1 = r_2 = \frac{1}{2};\ b = \frac{b_1 + b_2}{2};\ p_d = \frac{b_1}{b_1 + b_2} p_{id} + \frac{b_2}{b_1 + b_2} p_{gd}.$$

After lots of analysis and simulation experiments, Trelea emphasized on two sets of values for the parameters $a$ and $b$ which respectively correspond to PSO Trelea-I ($a=0.6, b=1.7$) and PSO Trelea-II ($a=0.729, b=1.494$).

*B. PSO with Clerc-Type 1 Constriction*

In order to constrain as well as control the velocities of the swarm particles, Clerc and Kennedy [22] suggested the use of a constriction factor in the basic PSO algorithm. The constriction factor is given by:

$$\chi = \begin{cases} \dfrac{2\kappa}{\phi - 2 + \sqrt{\phi^2 - 4\phi}} & \text{if } \phi \geq 4 \\ \kappa & \text{if } 0 < \phi < 4 \end{cases} \quad (15)$$

where, $0 < \kappa < 1$ and $\phi = b_1 + b_2$.

The calculated velocity through Eq. 11 is multiplied at each step with the constriction factor. This formulation is known as Clerc-Type1 PSO. The values $\phi = 4$ and $\kappa = 0.729$ are often found suitable in many practical applications.

## V. EMPIRICAL RESULTS AND DISCUSSIONS

Three time series with dominant seasonal patterns are used in our experiment. These are: (1) *Airline*: contains the number of monthly international airline passengers (in thousands) from January, 1949 to December, 1960, (2) *Red wine*: contains the monthly sales of red wine (in thousands of liters) in Australia from January, 1980 to December, 1995, (3) *Industry*: contains quarterly industrial observations, starting from the first quarter of 1977 and ending at the last quarter of 1992. The Airline and Red wine series are collected from the online Time Series Data Library (TSDL) [29], whereas the Industry series is the $219^{th}$ quarterly time series having ID N 863 of the M3 forecasting competition [1]. The description of these series is presented in Table I and their respective time plots are shown in Fig. 3.

TABLE I. DESCRIPTION OF THE THREE SEASONAL TIME SERIES

| Dataset | Total size | Training size (N) | s | $r_s$ | $r_{2s}$ | $2/\sqrt{N}$ |
|---------|------------|-------------------|---|-------|----------|--------------|
| Airline | 144 | 132 | 12 | 0.748 | 0.514 | 0.174 |
| Red wine | 187 | 168 | 12 | 0.777 | 0.658 | 0.154 |
| Industry | 64 | 48 | 4 | 0.888 | 0.775 | 0.289 |

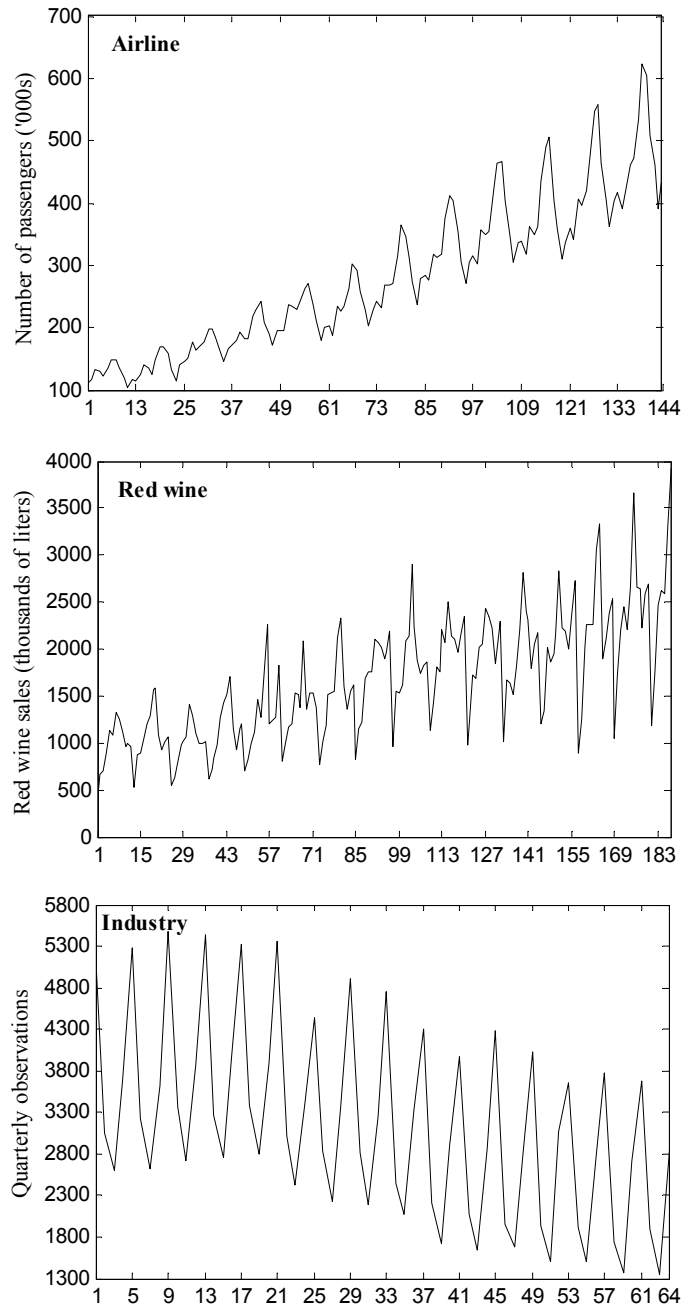

Fig. 3. Time plots of the three seasonal time series

Each time series shows strong seasonal fluctuations which is obvious from the time plots of Fig. 3 as well as from the rule of thumb, as discussed in Section II. The rule of thumb for each time series can be verified from Table I.

We fit the SARIMA, ANN and SVM models through the MATLAB software, whereas the HW models through the R environment [30]. The neural network toolbox [28] is used for BP training and the PSO toolbox of Birge [31] is used for implementing the three variants of the PSO algorithm. The observations of each time series are normalized to lie in the interval [0, 1] as follows:

$$y_i^{(\text{new})} = \frac{y_i - y^{(\min)}}{y^{(\max)} - y^{(\min)}} \quad (16)$$

$$\forall i = 1, 2, \ldots, N.$$

where, $\mathbf{Y} = \left[ y_1, y_2, \ldots, y_{N_{\text{train}}} \right]^T$ is the training dataset and $\mathbf{Y}^{(\text{new})} = \left[ y_1^{(\text{new})}, y_2^{(\text{new})}, \ldots, y_N^{(\text{new})} \right]^T$ is the normalized dataset, $y^{(\min)}$ and $y^{(\max)}$ respectively being the minimum and maximum values of the dataset $\mathbf{Y}$. After the forecasting experiment, all the normalized observations are again transformed back to their original values.

The SARIMA(0,1,1)×(0,1,1)$^s$ is found to be the suitable Box-Jenkins model for the three time series. The feedforward SANN structure for the Airline, Red wine and Industry series is determined to be 12×1×12, 12×2×12 and 4×3×4 respectively. It is observed that increasing the PSO swarm size beyond the range 24 to 30 often deteriorates the performance [20]. As such, the swarm size is kept to the minimum, i.e. 24 for all time series. The forecasting accuracies are evaluated through the two well-known error measures: Mean Absolute Error (MAE) and Mean Squared Error (MSE) [16, 20]. The obtained forecasting results are presented in Tables II, III and IV. In particular, the results for Red wine and Industry dataset are given in transformed scale (original MAE=MAE×10$^2$ and original MSE=MSE×10$^4$).

TABLE II. FORECASTING RESULTS OF THE THREE STATISTICAL MODELS

| Method | Airline | | Red Wine | | Industry | |
|---|---|---|---|---|---|---|
| | MAE | MSE | MAE | MSE | MAE | MSE |
| SARIMA | 17.08 | 411.75 | 2.42 | 9.01 | 1.53 | 4.84 |
| HW | 10.48 | 254.44 | 2.40 | 9.60 | 1.80 | 5.49 |
| SVM | 10.85 | 176.89 | 3.00 | 12.85 | 1.96 | 5.53 |

TABLE III. FORECASTING RESULTS OF THE SFANN MODELS

| Training algorithm | Airline | | Red Wine | | Industry | |
|---|---|---|---|---|---|---|
| | MAE | MSE | MAE | MSE | MAE | MSE |
| BP | 10.41 | 175.69 | 1.73 | 4.79 | 1.18 | 2.17 |
| PSO-Trelea1 | 9.12 | 150.41 | 1.47 | 3.62 | 1.05 | 1.65 |
| PSO-Trelea2 | 9.98 | 146.18 | 1.71 | 4.29 | 0.90 | 1.59 |
| PSO-Clerc | 9.50 | 142.12 | 1.44 | 3.57 | 1.03 | 1.72 |
| PSO-Average | 9.34 | 133.70 | 1.33 | 3.06 | 0.92 | 1.44 |
| PSO-Median | 9.16 | 138.34 | 1.30 | 2.98 | 0.93 | 1.49 |

TABLE IV. FORECASTING RESULTS OF THE SEANN MODELS

| Training algorithm | Airline | | Red Wine | | Industry | |
|---|---|---|---|---|---|---|
| | MAE | MSE | MAE | MSE | MAE | MSE |
| BP | 10.07 | 171.24 | 2.13 | 6.97 | 1.48 | 2.81 |
| PSO-Trelea1 | 9.85 | 161.76 | 1.94 | 5.57 | 0.95 | 1.40 |
| PSO-Trelea2 | 9.50 | 143.35 | 1.96 | 4.69 | 1.05 | 1.79 |
| PSO-Clerc | 9.84 | 161.96 | 2.02 | 5.93 | 1.01 | 1.65 |
| PSO-Average | 9.45 | 147.29 | 1.53 | 3.49 | 0.96 | 1.48 |
| PSO-Median | 9.50 | 144.90 | 1.84 | 4.76 | 0.92 | 1.38 |

A careful study of Tables II, III and IV reveals the following facts:
- The forecast errors obtained through the neural network models are notably less than those obtained through the traditional statistical models for all the three seasonal time series.
- Each of the three PSO variants achieves remarkably better forecasting accuracies, as compared to the standard BP algorithm for both feedforward as well as Elman SANN models.
- Among the three variants of the PSO algorithm, viz. Trelea-I, Trelea-II and Clerc-Type1, none can be declared as the best one.
- Overall better accuracies are obtained by combining the forecasting outputs of the three PSO based ANN models through simple average and median.

For visual illustration, the actual values (in solid line) and their corresponding forecasts (in dotted line) through PSO Trelea-I based SFANN models are presented in Fig. 4.

## VI. CONCLUSIONS

In this paper, an effort was made to assess the ability of PSO based neural networks to forecast seasonal time series. Three versions of the PSO algorithm were used to train two types of ANN models, viz. feedforward ANN (FANN) and Elman ANN (EANN). The three PSO variants are Trelea-I, Trelea-II and Clerc-Type1. Empirical analysis was conducted on three time series of which two had monthly and one had quarterly seasonal patterns. Obtained results demonstrated that each of the three PSO variants achieved reasonably better forecasting accuracies than the standard backpropagation (the Levenberg-Marquardt) algorithm for FANN as well as EANN models in terms of MAE and MSE. It was also observed that the issue of choosing the best PSO variant can be resolved by combining the forecasting outputs from all the three PSO based ANN models. Moreover, our study showed that both FANN and EANN provided remarkably better forecasting accuracies than the traditional SARIMA, HW and SVM methods for all three seasonal data. This study also supports the fact that ANNs are indeed capable of directly forecasting the inherent seasonal pattern without removing it from the time series. In future, the strength of PSO based ANN models can be further explored for other varieties of seasonal data.

ACKNOWLEDGMENT

The first author is grateful to the Council of Scientific and Industrial Research (CSIR) for the obtained financial support to carry out the present research work.

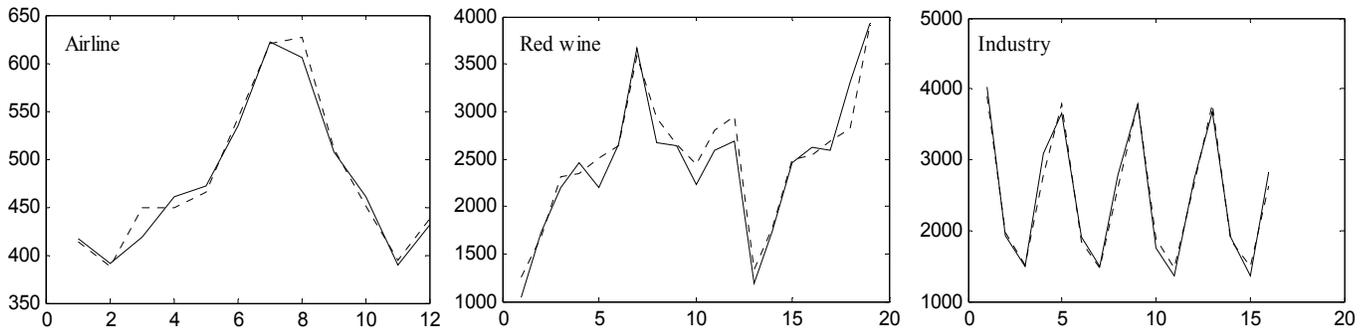

Fig. 4. Actual and forecasted observations of the three seasonal time series datasets